\documentclass[sigconf, review=False]{acmart}

\newcounter{mycounter}
\newcommand{\srcount}{\stepcounter{mycounter}\themycounter}

\usepackage{etoolbox}
\usepackage{booktabs}
\usepackage{subcaption}

\newtoggle{hide}
\newtoggle{show}
\newtoggle{CommentsMode}
\toggletrue{hide}
\newcommand{\nb}[1]{{\color{red}\small \iftoggle{hide}{}{\textbf{Nipun's comment:} #1}}}

\newcommand{\zp}[1]{{\color{blue}\small \iftoggle{hide}{}{\textbf{Zeel's comment:} #1}}}

\AtBeginDocument{%
  \providecommand\BibTeX{{%
    \normalfont B\kern-0.5em{\scshape i\kern-0.25em b}\kern-0.8em\TeX}}}

\setcopyright{acmlicensed}
\copyrightyear{2018}
\acmYear{2018}
\acmDOI{XXXXXXX.XXXXXXX}

\acmConference[Conference acronym 'XX]{Make sure to enter the correct
  conference title from your rights confirmation emai}{June 03--05,
  2018}{Woodstock, NY}
\acmISBN{978-1-4503-XXXX-X/18/06}

\begin{document}


\title[Scalable Methods for Brick Kiln Detection and Compliance Monitoring from Satellite Imagery]{Scalable Methods for Brick Kiln Detection and Compliance Monitoring from Satellite Imagery: A Deployment Case Study in India}

\author{Rishabh Mondal}
\authornote{Both authors contributed equally to this research.}
\email{rishabh.mondal@iitgn.ac.in}
\affiliation{%
  \institution{IIT Gandhinagar}
  \streetaddress{Palaj}
  \city{Gandhinagar}
  \state{Gujarat}
  \country{India}
  \postcode{382355}
}
\author{Zeel B Patel*}
\email{patel_zeel@iitgn.ac.in}
\affiliation{%
  \institution{IIT Gandhinagar}
  \streetaddress{Palaj}
  \city{Gandhinagar}
  \state{Gujarat}
  \country{India}
  \postcode{382355}
}

\author{Vannsh Jani}
\email{vannsh.jani@iitgn.ac.in}
\affiliation{%
  \institution{IIT Gandhinagar}
  \streetaddress{Palaj}
  \city{Gandhinagar}
  \state{Gujarat}
  \country{India}
  \postcode{382355}
}
\author{Nipun Batra}
\email{nipun.batra@iitgn.ac.in}

\affiliation{%
  \institution{IIT Gandhinagar}
  \streetaddress{Palaj}
  \city{Gandhinagar}
  \state{Gujarat}
  \country{India}
  \postcode{382355}
}



\begin{abstract}

Air pollution kills 7 million people annually. Brick manufacturing industry is the second largest consumer of coal contributing to 8\%-14\% of air pollution in Indo-Gangetic plain (highly populated tract of land in the Indian subcontinent). 
As brick kilns are an unorganized sector and present in large numbers, detecting policy violations such as distance from habitat is non-trivial.
Air quality and other domain experts rely on manual human annotation to maintain brick kiln inventory. Previous work
used computer vision based machine learning methods to detect brick kilns from satellite imagery but they are limited to certain geographies and labeling the data is laborious. In this paper, we propose a framework to deploy a scalable brick kiln detection system for large countries such as India and identify 7477 new brick kilns from 28 districts in 5 states in the Indo-Gangetic plain. 
We then showcase efficient ways to check policy violations such as high spatial density of kilns and abnormal increase over time in a region. We show that 90\% of brick kilns in Delhi-NCR violate a density-based policy. Our framework can be directly adopted by the governments across the world to automate the policy regulations around brick kilns.


\end{abstract}





\received{20 February 2007}
\received[revised]{12 March 2009}
\received[accepted]{5 June 2009}

\maketitle

\section{Introduction}



Air pollution kills seven million people worldwide, and 22\% of casualties are only from India~\citep{unep19}. 
Annual average PM$_{2.5}$ (Particulate matter of size $\le$ 2.5 $\mu$m) of India was 24 $\mu$g/m$^3$ in 2020, which is significantly higher than the annual WHO limit of 5 $\mu$g/m$^3$~\citep{guttikunda2022evolution}. Air quality researchers use physics-based simulators such as CAMx\footnote{\url{https://www.camx.com/}} to model the air quality~\citep{guttikunda2019air} using an inventory of major sources. Brick kilns are one such major source of pollution. They contribute to 8\%-14\% of air pollution in South Asia~\citep{worldbank2020}. Also, in South Asia, 144,000 units of brick kilns produce 0.94M tonnes of PM, 3.9M tonnes of CO, and 127M tonnes of $CO_{2}$ annually employing 15M workers~\citep{rajarathnam2014}.

\begin{figure}[h]
    \centering    \includegraphics[width=\linewidth]{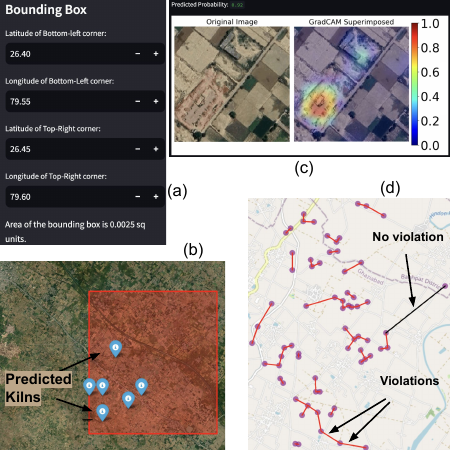}
    \caption{\href{https://brick-kilns-detector.streamlit.app/}{We have developed a Streamlit web App} to detect brick kilns in a) user-specified bounding box; b) predicted brick kilns by our model; c) GradCAM~\cite{selvaraju2016grad} run to validate that our model actually focuses on brick kiln to make a decision; d) 684 out of 762 brick kilns  (90\%) in a highly populated area called Delhi-NCR are within 1 km distance from another kiln, which is a violation of the government policy (violating distances are shown in red color).}
    \label{fig:intro}
\end{figure}

Monitoring these small, unregulated kilns using traditional survey methods is labor and resource-intensive and lacks scalability for maintaining the inventory due to their dynamic nature. Air quality and other domain experts who run physical models such as CAMx~\cite{camx}, leverage satellite imagery to detect these kilns with manual annotation. According to air quality experts, it takes around 6-8 hours to manually scan an area of size 3600 km$^2$.
Scaling this for a country like India would require more than 7000 hours of work. Often, brick kilns are installed at new locations, further making manual surveys hard to scale. Recent studies~\citep{lee2021scalable} have leveraged popular pretained CNN models like VGG16~\citep{huang2018densely}, ResNet~\citep{he2015deep} and EfficientNet~\cite{tan2020efficientnet} for transfer learning-based identification of brick kilns with imagery from a private satellite. However, such methods require extensive human annotation and expertise to curate a vast dataset.

Our paper deploys scalable methods for identifying brick kilns using publicly available satellite imagery. We collect over \textbf{7 million satellite images from Google Maps Static API} for this study. We use the model predicted locations in Bangladesh from a previous work~\cite{lee2021scalable} to create our initial dataset. We manually verify each of the predicted kilns to prepare the high quality ground truth dataset and evaluate the scalability on a highly populous region, Delhi-NCR, India. We do extensive ablation study on self-supervised learning techniques (JigSaw~\cite{noroozi2016unsupervised} and SimCLR~\cite{chen2020simple}) and transfer learning~\cite{lee2021scalable} to assess the performance under geography shift. We found that models pre-trained on large image datasets such as ImageNet are fine-tuned on brick kiln dataset, perform the best under geographical distribution shift. We then show that models can get confused with a similar looking images such as power plants. Including a small number of examples of power plants in training data, significantly reduced the number of power plants predicted as brick kilns. 

In the deployment stage, we identify the most populated districts from five states in Indo-Gangetic plain (the most polluted geographical region of India) and run our model on that for brick kiln detection. We manually verify and calculate ``precision'' of the model on those districts. Note that computing ``recall'' wouldn't be possible because it demands all positive examples to be known. We also study the violations of government policies around the brick kilns. Brick kilns needs to be at least 1 km distance apart from each other~\footnote{\url{https://cpcb.nic.in/uploads/Industry-Specific-Standards/Effluent/74-brick_kiln.pdf}}.

The \textbf{main findings} from our study are the following:

\begin{enumerate}
    \item Pre-trained and self-supervised learning based models do not perform well out-of-the-box on a different geography however they greatly improve with a small number of labeled images from the target region.
    \item We identify \textbf{7477} new brick kilns in Indo-Gangetic plain from 28 districts across 5 states analysing \textbf{276000 km$^2$ area}.
    \item We find that \textbf{90\%} of brick kilns in Delhi-NCR violate the government policy of minimum 1 km distance between brick kilns as shown in Figure~\ref{fig:intro}.
    \item The number of brick kilns in Delhi-NCR has increased by \textbf{15\%} in the previous twelve years.
\end{enumerate}


Finally, we have developed a web application\footnote{\url{https://brick-kilns-detector.streamlit.app/}} offering users an accessible and interactive interface for brick kiln detection in a given region of interest. Figure~\ref{fig:intro} shows our web application which takes in bounding boxes of the area of interest and detects the kilns present in the region while also showing Grad-CAM~\citep{Selvaraju_2019} visuals to highlight the focus area of the model. Our work is \textbf{fully reproducible}, and all details/code required to reproduce our work can be found at this repo~\footnote{\url{https://github.com/rishabh-mondal/kdd24_brick_kilns}}.









\section{Background and Related Work}
We now discuss the related work across: i) Machine Learning Applications on Satellite Imagery; and ii) Brick Kilns detection from satellite imagery.
\subsection{Machine Learning Applications on Satellite Imagery}

Satellite imagery serves as a low-cost and efficient tool with widespread applications, as demonstrated by studies~\cite{albert2017using} over conventional surveying techniques. Recent research has harnessed satellite imagery to predict diverse phenomena including poverty levels, cyclone intensity, flood occurrences, forest fires, and land cover. Previous work~\cite{rudner2019multi3net} proposes a method to generate segmentation maps of flooded buildings using publicly available images, thereby aiding first responders and government agencies in early disaster response efforts and providing crucial logistical and financial assistance to affected areas. Similarly, study~\cite{yang2021predicting} leverages remote-sensing satellite imagery to predict forest fires in Indonesia. The broad scope of satellite imagery extends to agriculture, as evidenced by research~\cite{dadsetan2021superpixels}, which utilizes aerial satellite imagery to segment agricultural areas based on nutritional deficiencies.

Various classes of machine learning tasks such as: object detection, classification, and regression tasks have been used across applications. A recent study~\cite{jean2016combining} employs Convolutional Neural Networks (CNNs) on high-resolution daytime images to quantify poverty using assets and expenditures across five African countries. Another study~\cite{ayush2020generating} combines object detection and regression techniques to predict consumption expenditure in Uganda from high-resolution satellite images, enabling the estimation of poverty levels. Further, a study~\cite{ayush2021efficient} utilizes a reinforcement learning-based CNN approach to acquire high-resolution images conditionally, facilitating the estimation of consumption expenditure in Uganda. ML techniques applied to satellite imagery also contribute to meteorological prediction, as demonstrated by research ~\cite{chen2021real}, which utilizes a GAN-CNN framework to estimate the intensity of tropical cyclones.

\subsection{Brick Kiln Detection from Satellite Imagery}

Machine learning techniques applied to satellite imagery have been utilized to identify brick kilns in South Asian regions. One study~\cite{nazir2020kiln} employs a gated neural network that decouples classification and object detection tasks to identify brick kilns. The study utilizes a deep learning architecture inspired by Inception-ResNet and the You Only Look Once (YOLO) object detector to locate brick kilns and evaluate it over 3300 km$^2$ area. In contrast, we evaluate our models on more than 100000 km$^2$ region (40x more region compared to Nazri et al.~\cite{nazir2020kiln}). Furthermore, recent studies have looked at alternate data sources beyond RGB imagery, such as heat signature data and gaseous emissions ~\cite{nazir2023detection}.\\
The scarcity of labeled data for training deep learning-based models has posed a challenge for brick kiln detection. Consequently, various approaches have been proposed to address this issue. A study by Misra et al. ~\cite{misra2019brick} employs a transfer learning-based approach to estimate brick kilns in North India using a limited dataset comprising 200 images per class. Most recently, Lee et al.~\cite{lee2021scalable} employed state-of-the-art machine learning models to predict over 7000 brick kilns in entire Bangladesh. We manually verified their predictions as our training data from Bangaldesh region. A recent report from United Nations Development Programme~\footnote{\url{https://www.undp.org/sites/g/files/zskgke326/files/2023-12/geoai_for_brick_kilns_v8_web_pages_002_0.pdf}} also employs YOLO algorithm to detect brick kilns around India but does not discuss the implementation level details. In contrast, we release our entire framework along with locations of detected brick kilns.

\section{Dataset}\label{sec:dataset}
In this section, we describe the dataset collection process and statistics of our data. Among multiple alternatives such as Google Earth Engine, Sentinel imagery and Google Maps Static (GMS) API, we found GMS API most suitable for our application due to lower rate limits and high resolution imagery processed by Google with production-level quality. 

\subsection{Google Maps Static (GMS) API}
GMS API combines multiple satellite products to provide satellite imagery at multiple zoom levels. It uses center of the image (in latitude and longitude) and provides maximum of $1200 \times 1200$ sized snapshot. Following the previous work~\cite{lee2021scalable}, we try to keep roughly 1 meter to pixel ratio for the images with zoom level 16 and scale=2 in the API. Note that by default the scale is 1 and thus zoom level 17 is needed to get 1 meter to pixel ratio images. With scale=2, \textbf{we reduced the number of images needed to be downloaded by 75\%} while getting the same resolution imagery from the GMS API. With a single API query, we get a $1200 \times 1200$ sized snapshot and center crop it to $1120 \times 1120$ size to create 25 $224 \times 224$ sized images from it ($5 \times 5$ grid)\nb{a lot of numbers: 1200, 1280, 25, 224. I'm a little confused. }. We experimentally found that, keeping $0.01$\nb{0.01 degree?} sliding window allows us to have a bit of overlap while ensuring full coverage of the entire region. With educational account in Google Cloud Platform, we could make $25000$ queries a day, with capacity of downloading 625k images per day per API key. We estimate that \textbf{country like India can be downloaded in 84 API days}. For example, if we have two API keys, we can download the entire dataset in 42 days. 

\subsection{Bangladesh dataset}
\begin{figure}[htbp]
  \centering
  \hspace{-2em}
  \begin{subfigure}[b]{0.2\textwidth}
    \centering
    \includegraphics[height=4.2cm]{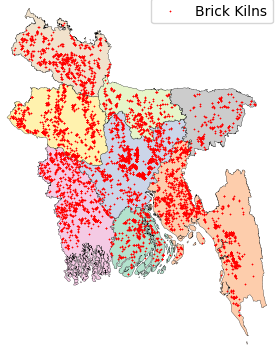}
    \caption{}
    \label{fig:subfig_a}
  \end{subfigure}
  \hspace{2em}
  \begin{subfigure}[b]{0.2\textwidth}
    \centering
    \includegraphics[height=4.2cm]{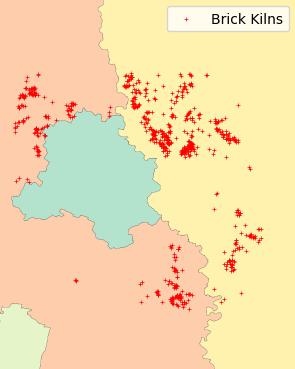}
    \caption{}
    \label{fig:subfig_b}
  \end{subfigure}
  \caption{Our brick kiln (labeled) datasets from (a) Bangladesh and (b) Delhi-NCR.}
  \label{fig:dataset}
\end{figure}

Previous work~\cite{lee2021scalable} employed state-of-the-art machine learning techniques to detect brick kilns and released over 7000 predicted brick kilns locations over Bangladesh. We mapped each of the location to its nearest two decimal variant to allow the sliding window of 0.01 and reduced them to 4500 locations. We manually labeled a total of 26000 images (1700 brick kilns) from this dataset. Figure~\ref{fig:dataset} (a) shows the locations of the labeled images in our dataset.


\begin{figure}[htbp]
  \centering
  \hspace{-2em}
  \begin{subfigure}[b]{0.2\textwidth}
    \centering
    \includegraphics[height=4.2cm]{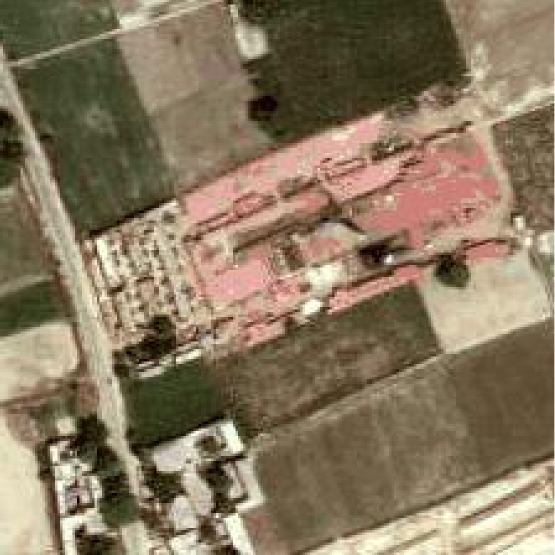}
    \caption{}
    \label{fig:subfig_a}
  \end{subfigure}
  \hspace{2em}
  \begin{subfigure}[b]{0.2\textwidth}
    \centering
    \includegraphics[height=4.2cm]{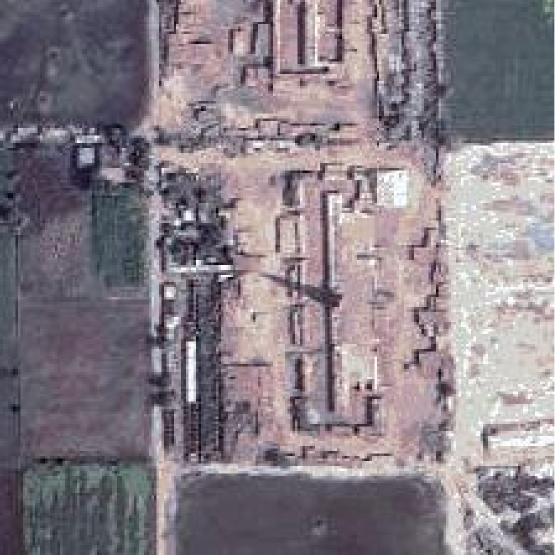}
    \caption{}
    \label{fig:subfig_b}
  \end{subfigure}
  \caption{Satellite images of a sample brick kiln from (a) Bangladesh and (b) India.}
  \label{fig:combined_images}
\end{figure}


\subsection{India dataset} 
\subsubsection{Delhi-NCR}
With guidance from an air quality domain expert \cite{guttikunda2019air}, we curated a dataset of 762 brick kilns across Delhi-NCR. We then selected 405 coordinates up to two decimal places and downloaded data according to specified settings in the previous section. We download a total of 10,025 images and manually label them resulting in 1,042 brick kiln images among all. We show the locations of labeled images in Figure~\ref{fig:dataset} (b).

\subsubsection{Other regions}\label{sec:data_other}
We identify the most populated districts in Indo-Gangetic plain and download the entire landscape of those districts for evaluation. In total, \textbf{we have downloaded the area of size approximately 276000 km$^2$ over 28 districts from 5 states}. 



\subsection{Data Labeling}
\begin{figure}[h]
\centering
\includegraphics[width=\linewidth]{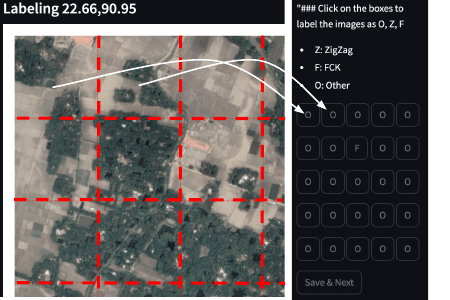}
    \caption{Brick kiln labeling App. Labeling individual images can be time consuming due to sparsity of brick kilns, thus, we develop an App to label a batch of 25 images ($5 \times 5$ grid) at once.}
    \label{fig:labeling_app}
\end{figure}

We have the model predicted brick kiln locations from the previous work~\cite{lee2021scalable} over Bangladesh but we have to manually verify them to create a ground truth dataset. Manually annotating images is time-consuming. Thus, we have created a data labelling app as shown in Figure~\ref{fig:labeling_app} to label 25 adjacent images at a time arranged in a $5 \times 5$ grid. It can significantly reduce the workload associated with labelling large datasets due to sparsity of brick kilns in the entire dataset. We have then also made a Moderator app that help moderators compare the labels from two annotators and break the ties. We ran the annotation process in team of 2 members and each member annotated the images in their respective teams independently. When the labels match between the annotators, we assume it to be correctly labeled image whereas for the conflicting labels a moderator's annotation is considered the final. We label an image as a brick kiln if 20\%-25\% of the brick kiln is visible within the image. \textbf{We labelled over 35000 images in the Bangladesh and the Delhi region in a short time span with help of these apps}.

\section{Approach}\label{sec:approach}
\subsection{Problem Formulation}
We formulate the problem of brick kiln detection as a one v/s rest (or binary) classification problem. Given an RGB image of size (224, 224, 3), the task is to classify it as either a brick kiln or not a brick kiln image. \zp{If we manage to add localization before deadline, we can add it here.}

\subsection{Convolutional Neural Network Models}
We pick some of the latest and most efficient models from the literature for our experiments, namely VGG~\citep{huang2018densely}, ResNet~\citep{he2015deep}, DenseNet~\citep{huang2018densely} and EfficientNet~\citep{tan2020efficientnet}. Here is a brief description of each of these models:
\begin{itemize}
    \item \textbf{VGG} uses uniform sized layers as opposed to increasing number of layers in typical CNN architectures and yet provides competitive performance. 
    \item \textbf{ResNet} uses residual connections to deal with the vanishing gradients problem.
    \item In \textbf{DenseNet}, a layer takes inputs from all of the preceding layers unlike vanilla CNN where each layer takes input only from its preceding layer.
    \item \textbf{EfficientNet} uses a compound scaling method that uniformly scales network depth, width, and resolution. This approach balances these dimensions to achieve optimal performance with fewer parameters. 
\end{itemize}

\subsection{Transfer Learning}
Transfer learning~\cite{huh2016makes} in computer vision implies to first train the model on a large dataset and then fine-tune further for the other usually smaller datasets. In this work, we use model weights trained on the ImageNet~\cite{deng2009imagenet} dataset which is a 1000 class dataset with over a million images. 

\subsection{Self Supervised Learning (SSL)}

Self-supervised learning (SSL) has emerged as a promising technique in computer vision applications. Unlike supervised learning, which requires labeled data, and unsupervised learning, which relies on unlabeled data, SSL offers a middle ground to learn the features automatically from unlabeled dataset on intelligently designed tasks. This approach helps us perform better on test data unseen by our models. In this research, we focus two self-supervised learning tasks: (\textbf{Sim}ple framework for \textbf{C}ontrastive \textbf{L}earning of visual \textbf{R}epresentations) SimCLR~\cite{chen2020simple} and Jigsaw~\cite{noroozi2016unsupervised}.

To the best of our knowledge, \textbf{we are the first to explore SSL for automatic brick kiln detection from satellite imagery}.

\subsubsection{SimCLR:} SSL has emerged as a pivotal approach in the realm of visual representation learning, exemplified by frameworks such as SimCLR. It mainly contains the following components:
\begin{itemize}
    \item A stochastic data augmentation module that generates two correlated views of the same data example, forming positive pairs crucial for learning.
    \item A flexible neural network encoder capable of extracting representation vectors from augmented data examples, offering versatility in network architecture selection.
    \item A compact neural network projection head responsible for mapping representations to a space where the contrastive loss function is applied. SimCLR utilizes a one-hidden-layer MLP with ReLU non-linearity for this purpose.
    \item The contrastive loss function, essential for the pretext task in SimCLR, employs the NT-Xent (Normalized Temperature-Scaled Cross-Entropy Loss) criterion, which facilitates the learning of discriminative representations. Mathematically, the NT-Xent loss function is defined as:\\
\begin{equation*}
\text{NT-Xent}(z_i, z_j) = -\log \frac{\exp(\text{sim}(z_i, z_j) / \tau)}{\sum_{k=1}^{2N} \mathbb{1}_{[k \neq i]} \exp(\text{sim}(z_i, z_k) / \tau)}
\end{equation*}\\
    where $z_i$ and $z_j$ are the representations of correlated views $x_i$ and $x_j$, $\text{sim}(z_i, z_j)$ is the cosine similarity between $z_i$ and $z_j$, $\tau$ is the temperature parameter, and $\mathbb{1}_{[k \neq i]}$ is the indicator function ensuring that $z_k$ is not the representation of the same sample as $z_i$.
    
\end{itemize} 

\subsubsection{JigSaw Method:} Jigsaw SSL~\cite{noroozi2016unsupervised} is another notable approach in visual representation learning. The procedure of Jigsaw SSL is as the following:
\begin{itemize}
    \item Each image is divided into $N \times N$ grid, to generate $N^2$ patches and $N^2!$ permutations per image. In cases where $N^2!$ is huge, $k$ permutations are chosen by maximizing hamming distance among the permutation orders.

    \item The SSL pre-training task is modeled as a classification task with \textit{k} target classes, where the model needs to predict the correct permutation number $0...k$. 

    \item Individual patches are passed in their permuted order through a shared weight network to get a representation. The representations are then concatenated in the same order to get a single representation and it is passed to a dense neural network classifier to predict the permutation number.

    \item After pre-training, the classifier can be replaced by other task specific networks (another classifier in case of classification) to fine-tune the model for down-stream tasks.
    
\end{itemize}

\section{Evaluation}\label{sec:evaluation}
In this section, we evaluate our models with focus to generalize on a new territory. We assume Bangladesh data as a comprehensive labeled dataset available beforehand and Delhi-NCR as a new region where we have limited or no labeled data available. We throughout show the results from the EfficientNet~\cite{tan2020efficientnet} model as it performs the best in all of our experiments.

\subsection{Evaluation metrics}

We use precision, recall, and F1 scores for the positive class as primary metrics for comparison. We assess performance based on the metrics of true positives ($TP$), false positives ($FP$), and false negatives ($FN$).

\subsubsection{Precision:} This metric quantifies the false positive rate i.e. lower the number of false positives, better the precision. It is defined as:
\begin{equation*}
\text{Precision} = \frac{TP}{TP+FP}
\end{equation*}

\subsubsection{Recall:} Recall quantifies the false negative rate i.e. lower the number of false negatives, better the recall. It is defined as:
\begin{equation*}
\text{Recall} = \frac{TP}{TP+FN}
\end{equation*}

\subsubsection{F1 Score:} The F1 score combines both precision and recall into a single measure. It is formulated as:
\begin{equation*}
\textit{F1} = 2 \times \frac{\text{Precision}\times\text{Recall}}{\text{Precision} + \text{Recall}}
\end{equation*}

\subsection{Experiments}
We conduct the following experiments during the evaluation
\nb{S can be replaced by Exp1 or Experiment 1. and so on..I think it is better to do that rather than leave any confusion}
\begin{itemize}
    \item \textbf{S1:} First we do a four-fold cross-validation on Bangladesh dataset to see the performance of the model on the known region (same geographical distribution). This is done to analyze the quantitative upper bound on the performance.
    \item \textbf{S2:} Then we train the model on Bangladesh dataset and predict directly on Delhi-NCR dataset to see the zero-shot performance.
    \item \textbf{S3:} In a few settings, we train the model only on Delhi-NCR to get a sense of compative performance with other settings.
    \item \textbf{S4:} Finally, we combine $X\%$ of labeled data from the target region to the training region to see the improvement in the performance with increase in labeled dataset.

\end{itemize}

\subsection{Experimental setup}

For \textbf{S1} setting, in $k^{th}$ split, the corresponding test split is used as a test dataset. We hold-out 50\% of the Delhi-NCR dataset as test dataset for settings \textbf{S2, S3} and \textbf{S4}.

\subsubsection{Vanilla setup}
In this setup, we initialize our model with default weight initialization (not pre-trained on any dataset) and then train them on the training dataset. 

\subsubsection{Transfer learning}
In this setup, we initialize the model with ImageNet weights and then fine-tune them further on the training dataset.

\subsubsection{Self-supervised learning based models}
In this setup, we first pre-train our model on the SSL specific tasks and then fine-tune them on the training dataset.


\subsection{Results and Analysis}


\begin{table}[h]
    \centering
    \small
    \resizebox{0.9\linewidth}{!}{%
    \begin{tabular}{llrrr}
    \toprule
        Pre-trained & Fold & Precision & Recall & F1 \\
    \midrule
    No & 1 & 0.86&0.80 &0.83 \\
    No & 2 &0.93 &0.64 &0.76 \\
    No & 3 & 0.86&0.75 &0.80 \\
    No & 4 &0.89 &0.67 &0.76 \\
    \midrule
     & Mean &0.885 & 0.715&0.787 \\
    \midrule
    \midrule
    ImageNet & 1  &0.92 &0.95 &0.94 \\
        ImageNet & 2  & 0.94&0.94 &0.94 \\
        ImageNet & 3  & 0.94&0.92 &0.93 \\
        ImageNet & 4  & 0.96&0.90 &0.93 \\
    \midrule
        & Mean &0.940 &0.927 &0.935 \\
    \bottomrule
    \end{tabular}
    }
    \vspace{5pt}
    \caption{Results of stratified four-fold cross validation on the Bangladesh dataset with EfficientNet model. Pre-trained model improve significantly over vanilla models with 0.15 increment in F1 score.}
    \label{tab:res_b_cross}
\end{table}
\begin{table}[htbp]
    \centering
    \renewcommand{\arrayrulewidth}{5pt}
    \resizebox{1\linewidth}{!}{
    \begin{tabular}{llllrrr}
    \toprule
    Sr. & SSL  & Pretrain & Fine-tune & Precision & Recall & F1 \\
    \midrule
    \midrule
     \srcount & No  & - & B &0.50 &0.35 &0.41 \\
     \srcount  & No  & - & (1\% D) &0.00 &0.00 &0.00 \\
     \srcount  & No  & - & (5\% D) & 0.38&0.11 &0.17 \\
     \srcount  & No  & - & (10\% D) & 0.26&0.28 &0.27 \\
     \srcount  & No  & - & (25\% D) & 0.38&0.23 &0.28 \\
     \srcount  & No  & - & (50\% D) & 0.80 & 0.47 & 0.59\\
     \srcount  & No  & - & B+(1\% D) & 0.59 & 0.30 & 0.40\\ 
     \srcount  & No  & - & B+(5\% D) & 0.54 & 0.39& 0.45 \\
     \srcount  & No  & - & B+(10\% D) &0.54 & 0.42&0.47 \\
     \srcount  & No  & - & B+(25\% D) & 0.65&0.39 &0.49 \\
     \srcount  & No  & - & B+(50\% D) &0.75 &0.43 &0.55 \\
    \midrule
    \midrule
     \srcount   & No & Imgnt & B & 0.87&0.66 &0.75 \\
     \srcount   & No & Imgnt & (1\% D) & 0.64&0.35 &0.46 \\
     \srcount   & No & Imgnt & (5\% D) & 0.77&0.80 &0.78 \\
     \srcount   & No & Imgnt & (10\% D) & 0.87&0.78 &0.82 \\
     \srcount   & No & Imgnt & (25\% D) &0.85 &0.89 &0.87 \\
     \srcount   & No & Imgnt & (50\% D) & 0.90&0.88 &0.89 \\
     \srcount   & No & Imgnt & B+(1\% D) & 0.88& 0.66&0.76 \\
     \srcount   & No & Imgnt & B+(5\% D) &0.90 &0.81 &0.85 \\
     \srcount   & No & Imgnt & B+(10\% D) & 0.92& 0.84&0.88 \\
     \srcount   & No & Imgnt & B+(25\% D) & 0.87& 0.89&0.88 \\
     \srcount   & No & Imgnt & B+(50\% D) & \textbf{0.94}&0.85 &\textbf{0.89}\\
    \midrule
    \midrule
       \srcount & Jigsaw  & B + D & B &0.55 &0.72 &0.63 \\
       \srcount & Jigsaw  & B + D& B+(1\% D) &0.80 &0.69 &0.74 \\
       \srcount & Jigsaw  & B + D & B+(5\% D) &0.84 & 0.75&0.79 \\
       \srcount & Jigsaw  & B + D & B+(10\% D) & 0.91&0.77 &0.83 \\
       \srcount & Jigsaw  & B + D & B+(25\% D) &0.89 &0.82 &0.86 \\
       \srcount & Jigsaw  & B + D & B+(50\% D) &0.85 &\textbf{0.90} &0.88 \\
    \midrule
    \midrule
       \srcount & SCLR  & B + D & B & 0.60 & 0.43 & 0.50\\
       \srcount & SCLR  & B + D  & B+(1\% D) & 0.72 & 0.41 & 0.52\\
       \srcount & SCLR  & B + D & B+(5\% D) & 0.70 & 0.65 & 0.61 \\
       \srcount & SCLR  & B + D & B+(10\% D) & 0.86 & 0.64 & 0.74 \\
       \srcount & SCLR  & B + D & B+(25\% D) & 0.86 & 0.71 & 0.78\\
       \srcount & SCLR  & B + D & B+(50\% D) & 0.81 & 0.88 & 0.84 \\
    \midrule   
    \bottomrule
    \end{tabular}
    }
    \caption{Performance of various models on different datasets. B = Bangladesh, D = Delhi. Imgnt=ImageNet. SCLR=SimCLR. Pre-trained models with a small number of images from the test data domain can significantly improve the overall performance (F1 score).}
    \label{tab:main_metrics}
\end{table}

\begin{table}[h]
    \centering
    \small
    \begin{tabular}{lrrr}
    \toprule
        model & Fine-Tune & FP & TN  \\
    \midrule
        Imgnt  &B+(50\%D) & 33 & 4705 \\
        Imgnt  &B+(50\%D)+(50\%P)&1 &4737 \\
        Jigsaw  & B+(50\%D)&158 &4580 \\
        Jigsaw  & B+(50\%D)+(50\%P)&4 &4734 \\
    \bottomrule
    \end{tabular}
    \caption{Result of ImageNet model and Jigsaw SSL model on power plant dataset. B = Bangladesh, D = Delhi, P = Power plant. Fine-tuning with small number of images which look alike brick kilns (such as powerplants) can help \nb{reduce false positives}increase the precision of the model.}
    \label{tab:with_powerplant}
\end{table}

\subsubsection{\textbf{S1} results}
Table~\ref{tab:res_b_cross} shows the results of the the EfficientNet~\cite{tan2020efficientnet} model with and without pretraining on the ImageNet dataset and fine-tuned on the Bangladesh dataset in stratified four-fold setting. We see an significant increment of 15\% \nb{not 15\% by by 0.15 frmo .. to ...}in F1-score with the pre-trained version of the model. It implies that pre-trained models can provide much better performance over the vanilla models in small data regime.\nb{also do mention you are able to reproduce and get comparable results to Lee et al.}

\subsubsection{\textbf{S2} results}
\nb{table not referred}
When we train the model on entire Bangladesh data and test on the held-out Delhi-NCR, precision is mildly decreased from $94\%$ to $87\%$ but recall is significantly decreased from $92\%$ to $66\%$. On the other hand, SSL based models (Row 23 and 29) do not perform at par with the pre-trained model (Row 12) but are doing better than vanilla models (Row 1). 

\subsubsection{\textbf{S3} and \textbf{S4} results}
When we train the models on a subset of Delhi-NCR data, we see a good improvement in \textbf{S4} over \textbf{S3} for $1\%, 5\%$ and $10\%$. We observe this improvement consistently for both ImageNet (Row 13 to 22) and SSL models (Row 24 to 28 and 30 to 34). For $25\%$ and $50\%$, the improvement is not substantial because models have seen a lot of data from the unknown region now and thus seeing Bangladesh data does not add much value. We see that both ImageNet and SSL models improve over the vanilla setting over all cases (Row 1 to 11).

Finally, we take (Row 22) as the best performing model to do the further predictions on the other districts.
\section{Deployment}\label{sec:deploy}
We now discuss the different aspects of deployment of the model. 
\subsection{Potentially Confused Class - Power Plants}
While analyzing the results of the various experiments, we observed that on some occasions, models were predicting power plant images as brick kilns. Both power plants and brick kilns typically have chimneys for venting out exhaust. Hence, we decided to test our models specifically on a power plant dataset consisting of 9475 satellite images of power plants. We used 50\% of the images as the test set and tested the performance of the best model. We found that the \textbf{model predicted 33 power plants as brick kilns out of 4738 power plant images}. To reduce the false positives corresponding to power plants, we further fine-tuned our model on the remaining 50\% of the power plant images (as 0 class samples) and tested the performance on the test set. We then found the model's false positives reduce: it \textbf{predicted only 1 power plant as a brick kiln out of 4738 power plant images}.

\subsection{Large Scale Brick Kiln Detection}
In this section, we discuss brick kiln detection performed with our final model. We first download the raw images from 28 districts across 5 states covering 276000 km$^2$ area as mentioned in Section~\ref{sec:data_other}. We sort the districts in each state by their population from high to low and choose top $k$ districts for the inference. We run the inference on these images and manually check the precision (mark images as True Positive or False Positive). We can not find recall because it requires identifying all positive images.
\nb{mention why we do not find recall} 
During this process, \textbf{we identify 7277 new brick kilns with 81.72\% precision} as described in Table~\ref{Geographical Predictive Scores across India}. We also find the population living within $X$ km distance of brick kilns and find that 7.65 million people live within 1 km of brick kilns, 11.73 million within 2 km and 61.56 million within 10 km. The population data was taken from UrbenEmissions.Info~\footnote{\url{https://urbanemissions.info/blog-pieces/india-gridded-population/}}


\zp{I am unsure if it is done correctly. Population within 2 km of kilns makes sense but what is kilns within 2 km of population?}

\subsection{Policy Violation Detection}
Central Pollution Control Board (CPCB), India has several guidelines to restrict unregulated growth of brick kilns. Some of the guidelines are:
\begin{enumerate}
    \item Minimum distance between two brick kilns should be 1 km.
    \item Brick kilns should be established at a minimum distance of 0.8 km from Schools, Hospitals, Courts, Government Offices, Habitation and Fruit orchards. 
    \item Brick kilns should be at minimum 0.5 km from River, Natural Water Resources, Dams, Ailetland and 0.2 km from State and National Highways and Railway Lines.
    \item New Brick Kiln units are prohibited in areas declared as Over Exploited or Semi-Critical (OCS) by CGWA.
    \item Brick kiln activity is not allowed within the eco-sensitive zone.
\end{enumerate}

\begin{figure}[h]
    \centering    \includegraphics[width=0.85\linewidth]{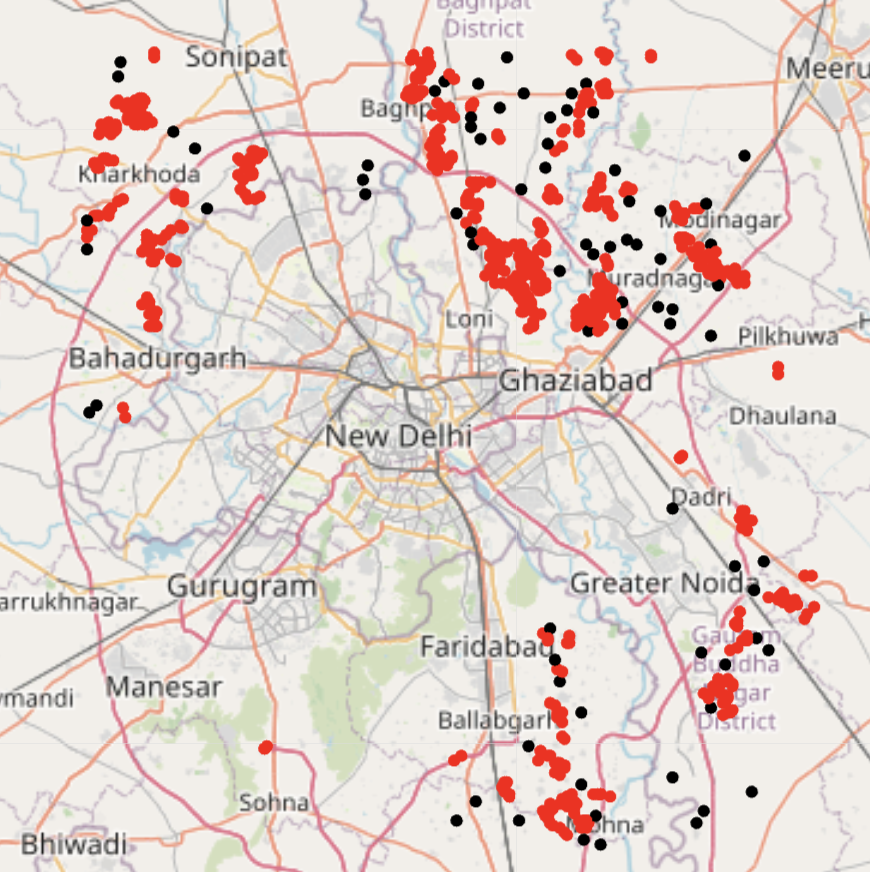}
    \caption{Brick kilns violating government policy of minimum distance 1 km between two brick kilns. 684 out of 762 (90\%) brick kilns violate this policy. \textcolor{red}{Red dots} show the kilns violating the policy and black dots show the kilns within the regulations.}
    \label{fig:violation}
\end{figure}

We calculate the pair-wise distances from our Delhi-NCR dataset and found that 684 out of 762
(90\%) brick kilns violate the first policy. Figure~\ref{fig:violation} shows the kilns violating the minimum distance policy. We believe that after our detection, if we have the other entity locations available, such that rivers and highways, we can find the number of kilns violating other regulations as well. Figure~\ref{kiln_years} illustrates an example of finding minimum distance from a river done manually which can be automated by including the river geolocation datasets.

\begin{figure}[h]
  \centering
  \includegraphics[width=1\columnwidth]{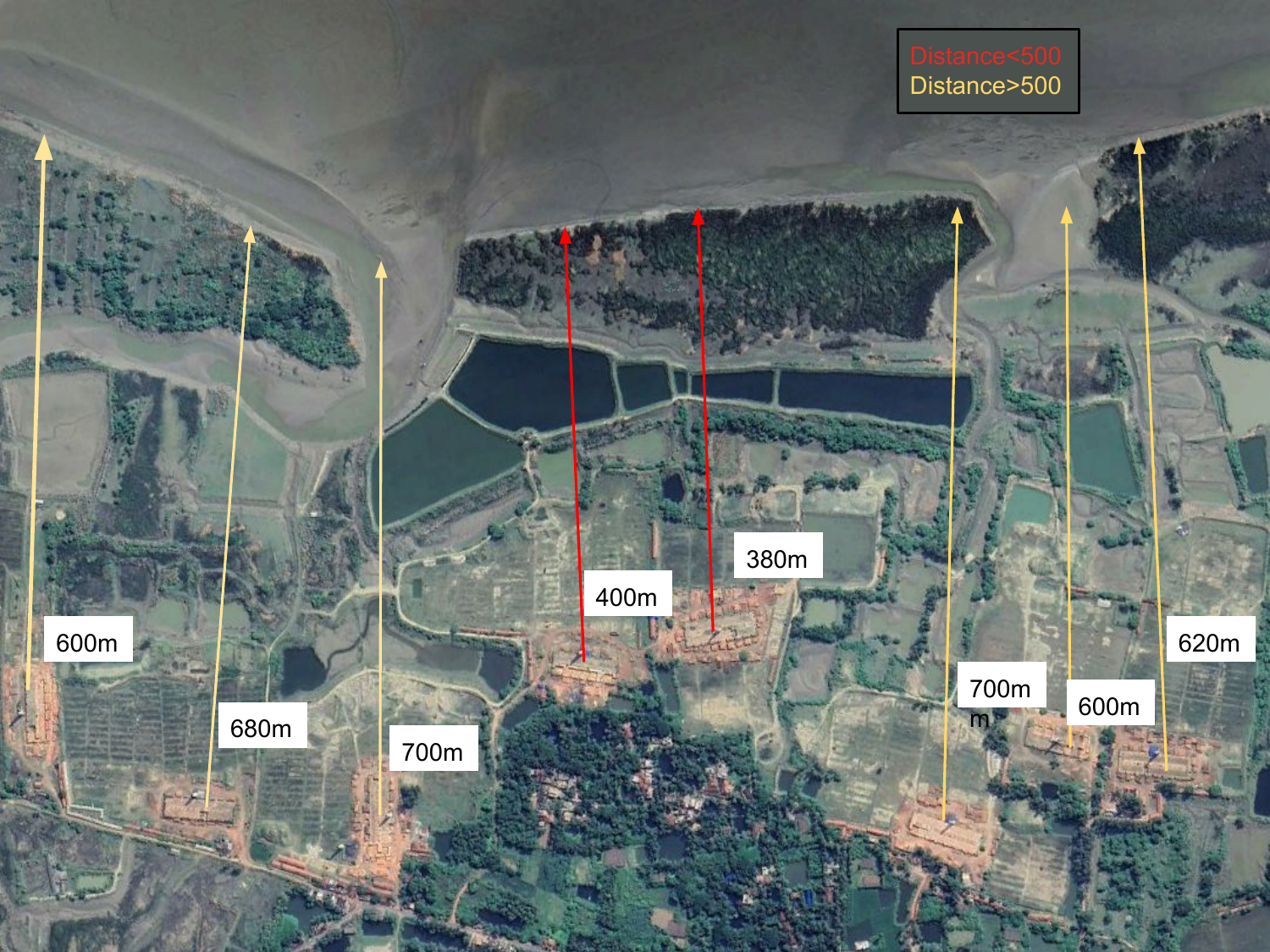}
  \caption{\nb{Figure legend should be distance of brick kiln to river and then we do not need Brick kilns = 1... also, the caption is not self contained}Assessment of Non-Compliance: Spatial Analysis of Brick Kilns in West Bengal State Relative to Ganga River Proximity. With access to rivers database, we can automatically identify all the kilns not complying with the government regulations.}
  \label{kiln_years}
\end{figure} 

\begin{table}[h]
\centering
\renewcommand{\arraystretch}{1.2}
\begin{tabular}{lcrrrr}
\toprule
\multicolumn{6}{c}{\textbf{Geographical Predictive Scores across India}} \\
\midrule
\textbf{State} & \textbf{District} & \textbf{TP} & \textbf{FP} & \textbf{P} & \textbf{Population} \\
\midrule
\midrule
 & Amritsar & {210} & {30} & {87\%} & 2,490,656 \\
& Ludhiana & 321 & 140 & 69\% & 3,498,739 \\
 & Jalandhar & 98 & 70 & 58\% & 2,193,590 \\
\textbf{\small{PB}} & Fazilka & 171 & 68 &  43\% & 107,000\\
  & Gurdaspur & 191 & 48 & 79\% & 2,298,323\\
   & Hoshiarpur & 169 & 139 & 55\% & 1,586,625\\
   & Pathankot & 95 & 23 & 85\% & 160,509 \\
   
\midrule
 & Jhajjar & {521} & {26} & {95\%} & {958,405}\\
 & Sonipat & 259 & 66 & 80\% & {1,450,001}\\
 & Jind & 156 & 30 & 87\% & {1,334,152}\\
 \textbf{\small{HR}}& Nuh & 176 & 25 & 87\% & {10,767}\\
 & Karnal & 165 & 56 & 74\%  & {1,505,324}\\
 & Rewari & 130 & 15 & 89\%  &{900,332}\\
 & Yamuna Nagar & 139 & 12 & 92\% &{1,214,205} \\

\midrule
& Kanpur  & {240} & {60} & {80\%} & {4,581,268}\\
 & Lucknow & 590 & 95 & 86\% &{4,589,838}\\
 & Baghpat & 577 & 115 & 83\% &{1,303,048}\\
\textbf{\small{UP}} & Jaunpur & 416 & 25 & 94\% &{4,494,204}\\
 & Kaushambi & 178 & 30 & 86\% &{1,599,596}\\
 & Kannauj & 144 & 36 & 80\% &{1,656,616}\\
\midrule
 & Patna & 283 & 145 & 66\% &{5,838,465}\\
 & Muzaffarpur & 394 & 38 & 91\% &{4,801,062}\\
 \textbf{\small{BR}}& Madhubani & 382 & 28 & 92\% &{4,487,379}\\
 & Darbhanga & 305 & 31 & 89\% &{3,937,385}\\
 & Saharsa & 132 & 41 & 76\% &{1,900,661}\\
\midrule
 & \small{East Medinipur}& {288} & {104} & 73\% &{5,095,875}\\
\textbf{\small{WB}} & \small{Malda} & 388 & 24 & 94\% &{3,988,845}\\
& \small{Birbhum}& 168 & 108 & 60\% &{3,502,404}\\
\midrule
& \textbf{Aggregate} & 7277 & 1628 & 81.72\% &{71,485,274}\\
\bottomrule
\end{tabular}
\vspace{5pt}
\caption{Prediction scores across various geographical regions in India utilizing the fine-tuned EfficientNetB0 model trained on data from Delhi, aiming to discern new brick kilns with heightened accuracy and efficiency. P=Precision. States: PB=Punjab, HR=Haryana, UP=Uttar Pradesh, BR=Bihar, WB=West Bengal.}
\label{Geographical Predictive Scores across India}
\end{table}

\begin{figure}[h]
  \centering
  \includegraphics[width=1\columnwidth]{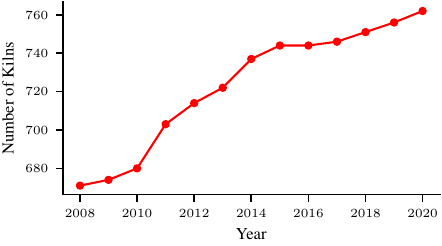}
  \caption{The number of brick kilns in Delhi-NCR has increased by 15\% in the previous twelve years. }
  \label{kiln_years}
\end{figure}


\subsection{Evolution of Kilns Over Time}
Finding the evolution of kilns over time can be helpful to understand the growth and subsequent effects (such as air pollution). Such studies can also help study effects of mitigation strategies or interventions. However, currently, GMS API does not provide a way to download the past images. We manually looked through historical imagery for the brick kiln sites that we verified in our study for the Delhi-NCR region in the Google Earth Engine Interface. We show the increase in brick kiln sites in the Delhi-NCR region in Figure~\ref{kiln_years}. We found that out of the ~760 brick kilns present today, less than 650 existed in 2008, and the numbers have increased by 15\% in the past 15 years. We emphasize that if satellite imagery is available, we can use our models (instead of manual inspection) to scale this detection over a large geographical space. In the future, we plan to train on models on lower-resolution satellite products that are available across a longer time period.
\section{LIMITATIONS AND FUTURE WORK}\label{sec:future}

We now address some limitations and propose future work.

\begin{itemize}
    \item In the current work, we have only looked at the image input. However, brick kilns often co-occur or are often clustered together owing to the presence of favourable soil conditions. Thus, leveraging location information within the model can help model localize better for that area. 
    \item Current literature is limited to using RGB imagery for brick kiln detection. We plan to use multi-spectral imagery to potentially improve the accuracy of predictions.
    \item In the current work, we have assumed each training example to be equally important. In the future, in an attempt to further scale our models, we plan to leverage active learning methods to query strategically. This may greatly reduce the labelling cost.
    \item In the current work, we have treated this problem as a classification task. In the future, we plan to study different problem variants such as i) object detection - to pinpoint the location of a brick kiln; and ii) segmentation - to accurately label the mask around the image. 
    \item In the future, we plan to use satellite imagery of lower resolutions such as Sentinel and Landsat products to further scale our proposed techniques. An additional advantage of these lower resolution imagery is the availability of historical data. 
    \item In the current work, we looked at a subset of the violations. In the future, we plan to automate the process of compliance checking for the remaining categories of violations.
\end{itemize}
\section{CONCLUSIONS}
In this paper, we presented a deployment of our methods for detecting brick kilns and their regulatory violations using satellite data and computer vision. We found that the proposed models exhibit a high precision and recall and thus can be used to aid for regulatory purposes and inventory management. Such scalable approaches can also maintain the inventory over time and thus be used to study effect of mitigation strategies. 
\bibliographystyle{ACM-Reference-Format}
\bibliography{sample-base}



\end{document}